\documentclass[conference]{IEEEtran}

\makeatletter
\def\@copyrightspace{\relax}
\makeatother

\usepackage{subfig}
\usepackage{graphicx}

\usepackage{program}
\usepackage{algorithm}
\usepackage{algpseudocode}

\usepackage{multirow}
\usepackage{boldline}
\usepackage{balance}

\usepackage[UKenglish]{babel}

\let\labelindent\relax
\usepackage{enumitem}

\begin{document}
\bstctlcite{IEEEexample:BSTcontrol}

\title{\fontsize{18}{18}\selectfont \textbf{AI Matrix: A Deep Learning Benchmark for Alibaba Data Centers}\vspace{-1.2cm}}

\author{Wei Zhang, Wei Wei, Lingjie Xu, Lingling Jin\\
        Alibaba Group\\
        E-mail: \{wz.ww, w.wei, lingjie.xu, l.jin\}@alibaba-inc.com\vspace{0cm}
        \and
        Cheng Li\\
        University of Illinois Urbana-Champaign\\
        E-mail: cli99@illinois.edu\vspace{0cm}}





\maketitle


\section{Introduction}
Alibaba has China's largest e-commerce platform. To support its diverse businesses, Alibaba has its own large-scale data centers providing the computing foundation for a wide variety of software applications. Among these applications, deep learning (DL) has been playing an important role in delivering services like image recognition, objection detection, text recognition, recommendation, and language processing. To build more efficient data centers that deliver higher performance for these DL applications, it is important to understand their computational needs and use that information to guide the design of future computing infrastructure. An effective way to achieve this is through benchmarks that can fully represent Alibaba's DL applications.

A number of DL benchmarks already exist or are under development, such as MLPerf \cite{mlperf}, DeepBench \cite{deepbench}, Training Benchmark for DNNs (TBD) \cite{dnnbench}, DAWNBench \cite{dawnbench}, Fathom \cite{fathom}, AI Benchmark \cite{aibench}, BenchIP \cite{benchip}, and synthetic benchmarks \cite{aimatrix}. However, these benchmarks are not good representations of Alibaba's DL workloads, for three reasons. First, some of these benchmarks, such as MLPerf, TBD, and DAWNBench,  are too general-purpose and target the most typical DL applications that are interesting to a broad range of users. Thus, they cannot represent the workloads that are specific to Alibaba's e-commerce environment. Second, many of the benchmarks, such as TBD, DAWNBench, Fathom, and BenchIP, have outdated and narrow collections that cannot catch up with the fast development of DL models and cover the diversity of DL applications in the e-commerce environment. Third, some of these benchmarks, such as DeepBench, AI Benchmark, BenchIP, and synthetic benchmarks, aim at testing specific tasks, e.g., the performance of specific operators, model layers, or Android systems. None of these aforementioned benchmarks satisfy the needs of fully characterizing the DL workloads in Alibaba's e-commerce environment, which motivates the development of Alibaba's in-house DL benchmark - AI Matrix.

AI Matrix results from a full investigation of the DL applications used inside Alibaba and aims to cover the typical DL applications that account for more than 90\% of the GPU usage in Alibaba data centers. This benchmark suite collects DL models that are either directly used or closely resemble the models used in the company's real e-commerce applications. It also collects the real e-commerce applications if no similar DL models are not available. Through the high coverage and close resemblance to real applications, AI Matrix fully represents the DL workloads on Alibaba data centers. The collected benchmarks mainly fall into three categories: computer vision, recommendation, and language processing, which consist of the most majority of DL applications in Alibaba.

AI Matrix serves a number of important purposes. It aids the selection of new hardware from outside vendors to build Alibaba's future data centers. Through performance analysis, it helps identify the bottleneck of the current DL software and hardware systems and provides guidance on improving application performance and designing future hardware. We believe that such a benchmark suite that fully characterizes the DL applications on China's largest e-commerce platform is of equal interest to the public, so we made the majority of the AI Matrix benchmarks open to the public, 17 out of 20, hoping it can benefit the hardware vendors, industrial and research organizations. More information of AI Matrix are available on the benchmark website \textit{https://aimatrix.ai/en-us/} and on GitHub \textit{https://github.com/alibaba/ai-matrix}.

\section{Collection of Models}
The model collection in AI Matrix mainly covers three categories: computer vision for image content understanding, recommendation for personalized feed, ranking, advertisement, etc, and language processing for translation, question and answer, searching, opinion analysis, etc. An overview of the model collection is shown in Table \ref{tab:overview}.

\subsection{Computer Vision}
Computer vision is an important application category on Alibaba e-commerce platform. With hundreds of millions of new pictures emerging every day, it is important to understand the content of these pictures. DL models are used to classify image content, detect objects, and recognize texts in images.

\textbf{Image Classification} classifies images into classes. The models collected for this task include GoogLeNet \cite{googlenet}, ResNet50 and ResNet101 \cite{resnet}, and DenseNet \cite{densenet}. These models are used as the backbone in many image classification applications in Alibaba.

\textbf{Object Detection} identifies specific regions that contain objects of interest and classifies them into classes. They use image classification models as backbone to extract features from the input image. The collected object detection models include SSD \cite{ssd}, DSSD \cite{dssd}, Mask RCNN \cite{maskrcnn}, Faster R-CNN \cite{fasterrcnn}, and Cascaded Pyramid Network (CPN) \cite{cpn}. The collected SSD and DSSD models include VGG \cite{vgg}, ResNet18, ResNet50, and ResNet101 as variants of the backbone network. SSD and DSSD are used in Alibaba's smart city application to optimize city traffic. The CPN is a model for human pose estimation, which is used in Taobao, an online shopping website.

\begin{table*}[t]
\centering
\begin{tabular}{ p{0.9cm}|p{1.2cm}|p{1.5cm}|p{1.2cm}|p{2.4cm}|p{0.6cm}|p{1.3cm}|p{1.4cm}|p{1cm}|p{1cm}|p{1.3cm} }
\hlineB{2}
\multicolumn{2}{c|}{\textbf{Category}} & \textbf{Model} & \textbf{Framework} & \textbf{Dataset} &\textbf{Batch Size} & \textbf{FLOPs} & \textbf{Memory Read (Bytes)} & \textbf{Arith. Intensity} & \textbf{Time/ Batch (s)} & \textbf{FLOPs/s} \\
\hlineB{2}
\multirow{10}{*}{} & \multirow{3}{*}{Classification} & GoogLeNet & Tensorflow \& Caffe & ImageNet \& Synthetic & 32 & $7.78\times10^{10}$	& $2.46\times10^{9}$ &	32 & 0.014 & $5.69\times10^{12}$ \\
\cline{3-11}
& & ResNet50 & Tensorflow \& Caffe & ImageNet \& Synthetic & 32 & $2.14\times10^{11}$ & $7.72\times10^{9}$ & 28 & 0.029 & $7.33\times10^{12}$ \\
\cline{3-11}
& & DenseNet121 & Tensorflow \& Caffe & ImageNet \& Synthetic & 32 & $9.19\times10^{10}$ &	$8.09\times10^{9}$ & 11 &0.029 & $3.21\times10^{12}$ \\
\cline{2-11}
& \multirow{5}{*}{Detection} & SSD & Tensorflow \& Caffe & PASCAL VOC & 16 & $1.13\times10^{12}$ & $2.69\times10^{10}$ & 42 & 0.250 & $4.53\times10^{12}$ \\
\cline{3-11}
Computer & & DSSD & Caffe & PASCAL VOC & 2 & $5.38\times10^{11}$ &	$9.52\times10^{09}$ & 57 & 0.091 & $5.92\times10^{12}$ \\
\cline{3-11}
Vision & & Mask RCNN & Tensorflow & MS COCO & 1 & $4.87\times10^{11}$ &	$1.65\times10^{10}$ &	30 & 0.138 & $3.52\times10^{12}$ \\
\cline{3-11}
& & Faster R-CNN & Caffe & PASCAL VOC & 1 & $3.91\times10^{11}$ &	$1.04\times10^{10}$ &	37 & 0.071 & $5.48\times10^{12}$ \\
\cline{3-11}
& & CPN & Tensorflow & MS COCO & 8 & $6.34\times10^{11}$	& $1.62\times10^{10}$ &	39 & 0.444 & $1.43\times10^{12}$ \\
\cline{2-11}
& \multirow{3}{*}{Text} & SegLink & Tensorflow & SynthText \& ICDAR15 & 8 & $2.75\times10^{12}$ &	$1.02\times10^{11}$ & 27 & 0.421 & $6.53\times10^{12}$ \\
\cline{3-11}
& & CRNN & Tensorflow & Synth 90k & 256 & $5.23\times10^{11}$ &	$8.48\times10^{9}$ &	62 & 0.086 & $6.09\times10^{12}$ \\
\hlineB{2}
\multicolumn{2}{c|}{\multirow{4}{*}{Recommendation}} & DIN & Tensorflow & Amazon Dataset & 512 & $4.01\times10^{09}$ &	$2.60\times10^{08}$ &	15 & 0.004 & $1.02\times10^{12}$ \\
\cline{3-11}
\multicolumn{2}{c|}{} & DIEN & Tensorflow & Amazon Dataset & 512 & $1.13\times10^{11}$ &	$5.10\times10^{09}$ &	22 & 0.051 & $2.24\times10^{12}$ \\
\cline{3-11}
\multicolumn{2}{c|}{} & Wide\&Deep & Tensorflow & Census Income  & 2048 & $1.30\times10^{08}$ &	$3.20\times10^{07}$ &	4 & 0.057 & $2.29\times10^{9}$ \\
\cline{3-11}
\multicolumn{2}{c|}{} & GCN & Tensorflow & Citation Network & 1 & $2.72\times10^{07}$ &	$2.15\times10^{06}$ &	13 & N/A & N/A \\
\cline{3-11}
\multicolumn{2}{c|}{} & NCF & Tensorflow & Movielens & 256 & $1.66\times10^{08}$ &	$3.34\times10^{06}$ &	50 & 0.001 & $1.57\times10^{11}$ \\
\hlineB{2}
\multicolumn{2}{c|}{\multirow{2}{*}{Language Processing}} & NMT & Tensorflow & WMT German-English & 128 & $2.59\times10^{12}$ &	$7.34\times10^{10}$ &	35 & 0.676 & $3.84\times10^{12}$ \\
\cline{3-11}
\multicolumn{2}{c|}{} & BERT & Tensorflow & Wikipedia, BooksCorpus, SQuAD & 2 & $5.39\times10^{11}$ &	$8.27\times10^{9}$ & 65 & 0.056 & $9.66\times10^{12}$ \\
\hlineB{2}
\end{tabular}
\caption{Overview of AI Matrix benchmarks. FLOPs is the number of FLOPs performed for executing one batch. Memory read is the total bytes read from DRAM to L2 cache for executing one batch. Arithmetic intensity is calculated using one batch.}
\vspace{-0.6cm}
\label{tab:overview}
\end{table*}

\textbf{Text Recognition} detects and extracts texts in images, which consists of a large fraction of the computer vision tasks in Alibaba. It has different challenges than the general objection detection. Text line bounding boxes usually have much larger aspect ratios and orientations than those of general objects. Faster R-CNN or SSD style detectors suffer from the difficulty of producing such boxes, due to its default box design. The collected text recognition models include SegLink \cite{seglink} and Convolutional Recurrent Neural Network (CRNN) \cite{crnn}. SegLink is used to identify the bounding boxes of texts. CRNN combines a convolutional network and a recurrent network to tackle the recognition of very long shaped texts.

\begin{figure}[t]
\centering
\vspace{-0.2cm}
\includegraphics[keepaspectratio, width=9cm]{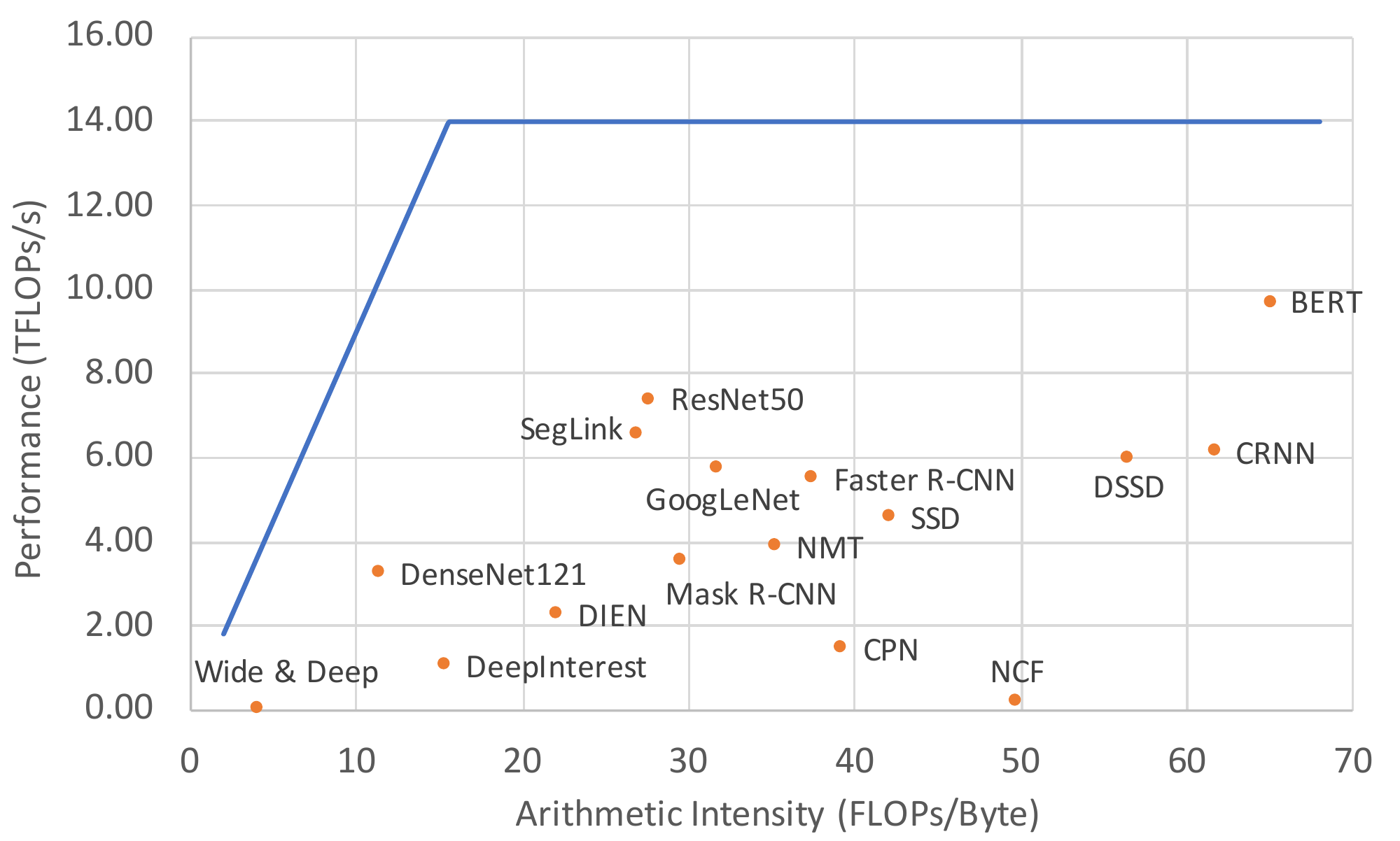}
\vspace{-0.6cm}
\caption{Roofline model of AI Matrix inference benchmarks on NVIDIA V100 GPU with peak FP32 performance of 14 TFLOPS and memory bandwidth of 900 GB/s. Precision is FP32 and no tensor core is used.}
\label{fig:arithmeticintensity}
\vspace{-0.8cm}
\end{figure}

\subsection{Recommendation}
Recommendation accounts for a large fraction of the DL applications in the e-commerce platform, with many use cases like advertisement, feed, and search. The collected recommendation models include Deep Interest Network (DIN) \cite{din}, Deep Interest Evolution Network (DIEN) \cite{dien}, Wide \& Deep \cite{widedeep}, Graph Convolutional Network (GCN) \cite{gcn}, and NCF \cite{ncf}. A commonly used scenario in advertisement is to predict the probability that a user will click a certain item if it is recommended to the user. DIEN is an important model used in Alibaba for this click-through rate prediction task. DIEN is an improved version of the DIN. GCN is a model used by Taobao to capture high-order similarities in users' billion-scale behavior sequences. These models combines a sparse embedding layer and a number of densely connected layers. The embedding layer transforms the sparse input of user features and history into a dense vector that can be handled by the densely connected layers. In real applications, the embedding size can be to hundreds of Gigabytes, while the fully-connected layers have a modest number of parameters.

\subsection{Language Processing}
Language processing is another important application category on Alibaba e-commerce platform and accomplishes tasks like translation, question \& answer, sentiment analysis, etc. The collected language models include Neural machine translation (NMT) \cite{gnmt} and Bidirectional Encoder Representations from Transformers (BERT) \cite{bert}. NMT has become the dominant approach to machine translation. It uses an encoder to encode the input sentence, and then uses a decoder to decode the encoding into the target output sentence. BERT is a new method of pre-training language representations which obtains state-of-the-art results on a wide array of language processing tasks. NMT and BERT are used as the backbone networks in many of the language processing tasks in Alibaba.

\section{Model Characteristics}
To understand the computation needs of AI Matrix benchmarks, it is important to look at their compute and memory intensities. Table \ref{tab:overview} shows the arithmetic intensity of these benchmarks for executing one batch. The majority of the benchmarks have high arithmetic intensity. Usually, the recommendation models, such as Wide \& Deep and DIN, have lower arithmetic intensity, because they mainly consist of fully-connected layers with high parameter count. On the other hand, the computer vision tasks like DSSD and CRNN have high arithmetic intensity, because they mainly consist of convolutional layers. The language processing models like NMT and BERT also have high arithmetic intensity because of the RNN layers used. The roofline model of AI Matrix benchmarks is shown in Figure \ref{fig:arithmeticintensity}. Surprisingly, even for those benchmarks with high arithmetic intensity, their performance are still far away from the roofline. This indicates that the compute resources on the GPU are not fully utilized. The reason why they are underutilized remains to be investigated, but one possibility is that the applications are not designed efficiently to fully utilize the GPU and its resources.

\section{Conclusion and Future Work}
The collection of models may be updated in the future based on the advancement of DL applications in Alibaba. In the short term, development will be done to support low-precision inference (FP16 and INT8), mixed-precision training (FP32 and FP16) on NVIDIA GPUs, training on single machine with multiple GPUs and on distributed machines.

\bibliographystyle{IEEEtran}
\bibliography{IEEEabrv,AIMatrix-SC19}

\begin{thebibliography}{10}
\providecommand{\url}[1]{#1}
\csname url@samestyle\endcsname
\providecommand{\newblock}{\relax}
\providecommand{\bibinfo}[2]{#2}
\providecommand{\BIBentrySTDinterwordspacing}{\spaceskip=0pt\relax}
\providecommand{\BIBentryALTinterwordstretchfactor}{4}
\providecommand{\BIBentryALTinterwordspacing}{\spaceskip=\fontdimen2\font plus
\BIBentryALTinterwordstretchfactor\fontdimen3\font minus
  \fontdimen4\font\relax}
\providecommand{\BIBforeignlanguage}[2]{{%
\expandafter\ifx\csname l@#1\endcsname\relax
\typeout{** WARNING: IEEEtran.bst: No hyphenation pattern has been}%
\typeout{** loaded for the language `#1'. Using the pattern for}%
\typeout{** the default language instead.}%
\else
\language=\csname l@#1\endcsname
\fi
#2}}
\providecommand{\BIBdecl}{\relax}
\BIBdecl

\bibitem{mlperf}
MLPerf, ``https://mlperf.org.''

\bibitem{deepbench}
DeepBench, ``https://github.com/baidu-research/deepbench.''

\bibitem{dnnbench}
H.~Zhu, M.~Akrout, B.~Zheng, A.~Pelegris, A.~Jayarajan, A.~Phanishayee,
  B.~Schroeder, and G.~Pekhimenko, ``Benchmarking and analyzing deep neural
  network training,'' in \emph{2018 IEEE International Symposium on Workload
  Characterization (IISWC)}, 2018.

\bibitem{dawnbench}
C.~Coleman, D.~Narayanan, D.~Kang, T.~Zhao, J.~Zhang, L.~Nardi, P.~Bailis,
  K.~Olukotun, C.~Re, and M.~Zaharia, ``Dawnbench: An end-to-end deep learning
  benchmark and competition,'' 2017.

\bibitem{fathom}
R.~Adolf, S.~Rama, B.~Reagen, G.-Y. Wei, and D.~Brooks, ``Fathom: Reference
  workloads for modern deep learning methods,'' in \emph{2016 IEEE
  International Symposium on Workload Characterization (IISWC)}, 2016.

\bibitem{aibench}
A.~Ignatov, R.~Timofte, W.~Chou, K.~Wang, M.~Wu, T.~Hartley, and L.~V. Gool,
  ``Ai benchmark: Running deep neural networks on android smartphones,'' in
  \emph{ECCV Workshops}, 2018.

\bibitem{benchip}
J.~Tao, Z.~Du, Q.~Guo, H.~Lan, L.~Zhang, S.~Zhou, L.-J. Xu, C.~Liu, H.-F. Liu,
  S.~Tang, A.~Rush, W.~Chen, S.~Liu, Y.~Chen, and T.~Chen, ``Benchip:
  Benchmarking intelligence processors,'' \emph{Journal of Computer Science and
  Technology}, 2017.

\bibitem{aimatrix}
W.~Wei, L.~Xu, L.~Jin, W.~Zhang, and T.~Zhang, ``{AI} matrix - synthetic
  benchmarks for {DNN},'' \emph{CoRR}, 2018.

\bibitem{googlenet}
C.~Szegedy, W.~Liu, Y.~Jia, P.~Sermanet, S.~Reed, D.~Anguelov, D.~Erhan,
  V.~Vanhoucke, and A.~Rabinovich, ``Going deeper with convolutions,'' in
  \emph{Computer Vision and Pattern Recognition (CVPR)}, 2015.

\bibitem{resnet}
K.~He, X.~Zhang, S.~Ren, and J.~Sun, ``Deep residual learning for image
  recognition,'' \emph{2016 IEEE Conference on Computer Vision and Pattern
  Recognition (CVPR)}, 2015.

\bibitem{densenet}
G.~Huang, Z.~Liu, and K.~Q. Weinberger, ``Densely connected convolutional
  networks,'' \emph{2017 IEEE Conference on Computer Vision and Pattern
  Recognition (CVPR)}, 2016.

\bibitem{ssd}
W.~Liu, D.~Anguelov, D.~Erhan, C.~Szegedy, S.~E. Reed, C.-Y. Fu, and A.~C.
  Berg, ``Ssd: Single shot multibox detector,'' in \emph{ECCV}, 2016.

\bibitem{dssd}
C.-Y. Fu, W.~Liu, A.~Ranga, A.~Tyagi, and A.~C. Berg, ``Dssd : Deconvolutional
  single shot detector,'' \emph{ArXiv}, 2017.

\bibitem{maskrcnn}
K.~He, G.~Gkioxari, P.~Doll{\'a}r, and R.~B. Girshick, ``Mask r-cnn,''
  \emph{2017 IEEE International Conference on Computer Vision (ICCV)}, 2017.

\bibitem{fasterrcnn}
S.~Ren, K.~He, R.~Girshick, and J.~Sun, ``Faster r-cnn: Towards real-time
  object detection with region proposal networks,'' in \emph{Proceedings of the
  28th International Conference on Neural Information Processing Systems -
  Volume 1}, 2015.

\bibitem{cpn}
Y.~Chen, Z.~Wang, Y.~Peng, Z.~Zhang, G.~Yu, and J.~Sun, ``Cascaded pyramid
  network for multi-person pose estimation,'' \emph{2018 IEEE/CVF Conference on
  Computer Vision and Pattern Recognition}, 2017.

\bibitem{vgg}
K.~Simonyan and A.~Zisserman, ``Very deep convolutional networks for
  large-scale image recognition,'' \emph{CoRR}, 2014.

\bibitem{seglink}
B.~Shi, X.~Bai, and S.~J. Belongie, ``Detecting oriented text in natural images
  by linking segments,'' \emph{2017 IEEE Conference on Computer Vision and
  Pattern Recognition (CVPR)}, 2017.

\bibitem{crnn}
B.~Shi, X.~Bai, and C.~Yao, ``An end-to-end trainable neural network for
  image-based sequence recognition and its application to scene text
  recognition,'' \emph{IEEE Transactions on Pattern Analysis and Machine
  Intelligence}, 2015.

\bibitem{din}
G.~Zhou, X.~Zhu, C.~Song, Y.~Fan, H.~Zhu, X.~Ma, Y.~Yan, J.~Jin, H.~Li, and
  K.~Gai, ``Deep interest network for click-through rate prediction,'' in
  \emph{Proceedings of the 24th ACM SIGKDD International Conference on
  Knowledge Discovery \& Data Mining}, 2018.

\bibitem{dien}
G.~Zhou, N.~Mou, Y.~Fan, Q.~Pi, W.~Bian, C.~Zhou, X.~Zhu, and K.~Gai, ``Deep
  interest evolution network for click-through rate prediction,'' \emph{ArXiv},
  2018.

\bibitem{widedeep}
H.-T. Cheng, L.~Koc, J.~Harmsen, T.~Shaked, T.~Chandra, H.~Aradhye,
  G.~Anderson, G.~Corrado, W.~Chai, M.~Ispir, R.~Anil, Z.~Haque, L.~Hong,
  V.~Jain, X.~Liu, and H.~Shah, ``Wide \& deep learning for recommender
  systems,'' in \emph{Proceedings of the 1st Workshop on Deep Learning for
  Recommender Systems}, 2016.

\bibitem{gcn}
T.~N. Kipf and M.~Welling, ``Semi-supervised classification with graph
  convolutional networks,'' \emph{ArXiv}, 2016.

\bibitem{ncf}
X.~He, L.~Liao, H.~Zhang, L.~Nie, X.~Hu, and T.-S. Chua, ``Neural collaborative
  filtering,'' \emph{ArXiv}, 2017.

\bibitem{gnmt}
Y.~Wu, M.~Schuster, Z.~Chen, Q.~V. Le, M.~Norouzi, W.~Macherey, M.~Krikun,
  Y.~Cao, Q.~Gao, K.~Macherey, J.~Klingner, A.~Shah, M.~Johnson, X.~Liu,
  L.~Kaiser, S.~Gouws, Y.~Kato, T.~Kudo, H.~Kazawa, K.~Stevens, G.~Kurian,
  N.~Patil, W.~Wang, C.~Young, J.~Smith, J.~Riesa, A.~Rudnick, O.~Vinyals,
  G.~S. Corrado, M.~Hughes, and J.~Dean, ``Google's neural machine translation
  system: Bridging the gap between human and machine translation,''
  \emph{ArXiv}, 2016.

\bibitem{bert}
J.~Devlin, M.~Chang, K.~Lee, and K.~Toutanova, ``{BERT:} pre-training of deep
  bidirectional transformers for language understanding,'' \emph{CoRR}, 2018.

\end{thebibliography}

\end{document}